\documentclass[sigconf]{aamas}

\usepackage{balance} 
\usepackage{enumitem}
\usepackage{hyperref}
\usepackage{siunitx}
\usepackage{tcolorbox} 
\usepackage{framed} 

\def\our{\texorpdfstring{\textsc{PyVRP}$^{+}$}{PyVRP+}}

\usepackage{newfloat}
\usepackage{listings}
\DeclareCaptionStyle{ruled}{labelfont=normalfont,labelsep=colon,strut=off} 
\floatstyle{ruled}
\newfloat{listing}{tb}{lst}{}
\floatname{listing}{Listing}

\doi{}

\makeatletter
\gdef\@copyrightpermission{
  \begin{minipage}{0.2\columnwidth}
   \href{https://creativecommons.org/licenses/by/4.0/}{\includegraphics[width=0.90\textwidth]{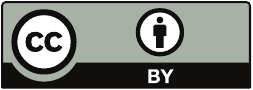}}
  \end{minipage}\hfill
  \begin{minipage}{0.8\columnwidth}
   \href{https://creativecommons.org/licenses/by/4.0/}{This work is licensed under a Creative Commons Attribution International 4.0 License.}
  \end{minipage}
  \vspace{5pt}
}
\makeatother

\setcopyright{ifaamas}
\acmConference[AAMAS '26]{Proc.\@ of the 25th International Conference
on Autonomous Agents and Multiagent Systems (AAMAS 2026)}{May 25 -- 29, 2026}
{Paphos, Cyprus}{C.~Amato, L.~Dennis, V.~Mascardi, J.~Thangarajah (eds.)}
\copyrightyear{2026}
\acmYear{2026}
\acmDOI{}
\acmPrice{}
\acmISBN{}

\title[\our: LLM-Driven Metacognitive Heuristic Evolution for Hybrid Genetic Search in Vehicle Routing Problems]{\our: LLM-Driven Metacognitive Heuristic Evolution for Hybrid Genetic Search in Vehicle Routing Problems}

\author{Manuj Malik*}
\affiliation{
  \institution{Singapore Management University}
  \city{Singapore}
  \country{Singapore}}
\email{manujm@smu.edu.sg}

\author{Jianan Zhou*}\thanks{$^*$ Corresponding authors.}
\affiliation{
  \institution{Nanyang Technological University}
  \city{Singapore}
  \country{Singapore}}
\email{jianan004@e.ntu.edu.sg}

\author{Shashank Reddy Chirra}
\affiliation{
  \institution{University of Oxford}
  \city{Oxford}
  \country{England}}
\email{shashank@robots.ox.ac.uk}

\author{Zhiguang Cao}
\affiliation{
  \institution{Singapore Management University}
  \city{Singapore}
  \country{Singapore}}
\email{zgcao@smu.edu.sg}

\begin{abstract}
Designing high-performing metaheuristics for NP-hard combinatorial optimization problems, such as the Vehicle Routing Problem (VRP), remains a significant challenge, often requiring extensive domain expertise and manual tuning. Recent advances have demonstrated the potential of large language models (LLMs) to automate this process through evolutionary search. 
However, existing methods are largely reactive, relying on immediate performance feedback to guide what are essentially black-box code mutations. Our work departs from this paradigm by introducing Metacognitive Evolutionary Programming (MEP), a framework that elevates the LLM to a strategic discovery agent. Instead of merely reacting to performance scores, MEP compels the LLM to engage in a structured Reason-Act-Reflect cycle, forcing it to explicitly diagnose failures, formulate design hypotheses, and implement solutions grounded in pre-supplied domain knowledge. By applying MEP to evolve core components of the state-of-the-art Hybrid Genetic Search (HGS) algorithm, we discover novel heuristics that significantly outperform the original baseline. 
By steering the LLM to reason strategically about the exploration-exploitation trade-off, our approach discovers more effective and efficient heuristics applicable across a wide spectrum of VRP variants. Our results show that MEP discovers heuristics that yield significant performance gains over the original HGS baseline, improving solution quality by up to 2.70\% and reducing runtime by over 45\% on challenging VRP variants.
\end{abstract}


\newcommand{\BibTeX}{\rm B\kern-.05em{\sc i\kern-.025em b}\kern-.08em\TeX}

\begin{document}


\pagestyle{fancy}
\fancyhead{}


\maketitle 

\vspace{-3mm}
\section{Introduction}

Vehicle routing problems (VRPs) represent the foundation of combinatorial optimization (CO) with profound implications for logistics, supply chain management, and transportation. Given their NP-hard nature, exact methods are intractable for large-scale, real-world instances, leading to the widespread adoption of metaheuristics. These high-level algorithmic frameworks are designed to efficiently explore the solution space for near-optimal solutions. Among these, Hybrid Genetic Search (HGS) has emerged as a State-Of-The-Art (SOTA) approach, demonstrating exceptional performance across numerous VRP variants \cite{vidal2022hybrid}. Despite its effectiveness, HGS remains heavily dependent on human-crafted algorithmic components. Like many metaheuristics, its performance hinges on carefully tuned internal mechanisms, such as parent selection, survivor selection, and penalty adjustment, which are designed through tedious trial and error and require significant domain expertise. 

Recently, Large Language Models (LLMs) have emerged as a promising tool for automatic heuristic design and scientific discovery \cite{novikov2025alphaevolve}. By leveraging their ability to reason over code, data, and mathematical structure, LLMs have the potential to surpass human-engineered solutions across a wide range of domains. For instance, FunSearch \cite{romera2024mathematical} integrated LLMs into an evolutionary framework to discover novel solutions to mathematical problems, while AlphaEvolve \cite{novikov2025alphaevolve} extended this paradigm to generate strategies that outperform human-designed algorithms in domains such as matrix multiplication, kernel engineering, and beyond. Beyond raw performance, LLM-driven evolution condenses the costly manual design cycle—often requiring weeks of domain expert effort—into a manageable automated process.

Similar progress has also been made in the domain of CO. Evolution of Heuristics (EoH) \cite{liu2024evolution} demonstrated that LLMs can effectively automate heuristic design within an evolutionary loop by co-evolving solutions at two levels of abstraction: high-level thoughts expressed in natural language and their corresponding executable code. 
ReEvo \cite{ye2024reevo} further leveraged ``verbal gradients'' derived from performance feedback to iteratively refine heuristics.

While promising, these approaches mainly focus on simple heuristics for standard CO problems such as the traveling salesman problem (TSP), and advancing traditional solvers beyond human capability remains a challenge. While these pioneering works successfully demonstrate the potential of LLM-driven evolution, they often treat the LLM as a reactive code generator, relying predominantly on immediate performance feedback to steer the search. This paper argues that such an approach underutilizes the reasoning capabilities of modern LLMs. Our core contribution is therefore not only the use of an LLM for evolution but also the introduction of Metacognitive Evolutionary Programming (MEP), a framework that shifts the paradigm from reactive mutation to proactive, structured discovery. MEP elevates the LLM from a simple code mutator to a strategic discovery agent by embedding two key innovations: a \emph{Domain-Aware Initialization} phase that grounds the model in established strategic principles, and a mandatory \emph{Reason-Act-Reflect} cycle that compels the LLM to explicitly diagnose failures, formulate testable design hypotheses, and critically self-assess its own output. This metacognitive scaffolding is the primary driver of our results, enabling the discovery of heuristics that not only perform high but also advance SOTA in VRP solving with minimal human intervention.



By framing the evolutionary search as a hypothesis-driven discovery process, our aim is to show that heuristic evolution can systematically uncover novel heuristics that outperform their human-designed counterparts. Our work builds on the framework introduced by \cite{bai2025mp}, which models human-like metacognitive reasoning. We structure our methodology in two key phases:

\begin{itemize}
    \item \textbf{Domain-Aware Initialization}: We equip the LLM with domain knowledge via three critical components: common pitfalls ($K_p$) that undermine the target component, proven mitigation strategies ($K_s$) to address these pitfalls, and problem-specific traps ($K_t$) unique to VRP solving. This knowledge foundation primes the LLM with the necessary domain understanding before the evolutionary search begins.
    \item \textbf{Reason-Act-Reflect Cycle}: Each generation in the evolutionary loop of MEP follows a structured cognitive cycle. The LLM is prompted to: 1) \emph{Reason}: Explicitly diagnose the shortcomings of parent heuristics. 2) \emph{Act}: Formulate a clear design hypothesis and implement it as a new code-based heuristic. 3) \emph{Reflect}: Assess its own generated code by articulating its rationale and potential limitations directly within the code's documentation.
 
\end{itemize}

We apply MEP to the challenging task of discovering high-performance components for the HGS algorithm, using the open-source PyVRP library \cite{wouda2024pyvrp} as our testbed. The discovered collection of high-performing heuristics, which we term \our, demonstrates that a structured, hypothesis-driven evolution can produce novel HGS variants that advance the state-of-the-art in VRP solving. The results show that MEP is capable of evolving heuristics that deliver performance gains of up to \textbf{2.70\%}, while simultaneously reducing runtime by over \textbf{45\%} on challenging VRP variants.
These improvements underscore MEP's potential as a compelling alternative to hand-crafted HGS components and represent an encouraging step toward automated algorithm discovery in CO.


\section{Related Work}

Recent research has increasingly adopted LLMs to automate the design of heuristics for CO. Pioneering this direction, FunSearch \cite{romera2024mathematical} demonstrated that LLMs could effectively generate core heuristic components, circumventing the need for manual redesign of entire algorithms. Similarly, \citet{liu2023algorithm} employed LLMs to generate key evolutionary algorithm components, such as crossover and mutation operators. Further advancing this line of work, the Evolution of Heuristics (EoH) framework \cite{liu2024evolution} introduced a method for iteratively refining heuristic seed functions through evolutionary computation.
Most of these approaches have adopted evolutionary frameworks, integrating advanced techniques to enhance their efficacy. These techniques include reflective feedback mechanisms \cite{ye2024reevo}, predictive modeling for performance estimation \cite{wu2025efficient}, and strategies that encourage heuristic diversity \cite{dat2025hsevo}. This paradigm has successfully extended beyond single-objective scenarios, addressing multi-objective optimization problems \cite{yao2025multi} and diverse CO domains including SAT \cite{sun2024autosat}, MILP \cite{ye2025large}, and dynamic Job Shop Scheduling Problems (JSSP) \cite{huang2026automatic}.
Moving beyond single-agent approaches, recent studies \cite{yangheuragenix, sun2024autosat} have explored multi-agent frameworks, achieving enhanced performance through collaborative heuristic generation. To overcome limitations associated with population-based methods, \citet{zheng2025monte} introduced a Monte Carlo Tree Search (MCTS) strategy, integrating LLM-generated heuristics to boost exploration efficiency.
More recent advancements focus on reducing manual intervention. \citet{thach2025redahd} proposed a novel end-to-end framework based on problem reduction, which enables heuristic methods to function independently of predefined general algorithmic templates. Similarly, \citet{shi2025generalizable} developed a meta-optimization framework that systematically discovers robust optimizers for a range of heuristic design tasks, promoting broader exploration and operating at a higher abstraction level.
Furthermore, \citet{vsurinaalgorithm} suggested integrating evolutionary heuristic generation with reinforcement learning-based fine-tuning, shifting LLM usage from static heuristic generation towards dynamic refinement of both heuristics and the underlying models. 
However, these approaches primarily focus on evolving simple heuristics or individual components thereof, rather than tackling sophisticated and well-established traditional solvers. Most existing work targets constructive heuristics for basic problem variants or designs standalone neural policies that bypass traditional optimization entirely \cite{bengio2021machine,cappart2023combinatorial}. Concurrently, LLMs have also been explored as end-to-end CO solvers \cite{jiang2025large}, for refining HGS via RL-finetuned LLMs \cite{zhu2025refining}, and within agentic frameworks for complex VRPs \cite{zhang2025agentic}. In this paper, we bridge this gap by demonstrating that \emph{LLM-guided evolution can enhance the performance of a SOTA solver like HGS across multiple challenging VRP variants, moving beyond proof-of-concept demonstrations to practical algorithmic improvements.} For a comprehensive review of the use of LLMs in CO, we direct readers to \cite{da2025large}.

LLM-guided scientific discovery is rapidly expanding beyond combinatorial optimization. For instance, \citet{chen2023evoprompting} apply LLMs to Neural Architecture Search, \citet{goldie2025how} use them to discover novel loss functions and optimizers for reinforcement learning, and \citet{nadimpalli2025evolving} explore their use in evolving activation functions. While most of these methods focus on evolving function-level programs, \citet{novikov2025alphaevolve} scale this approach to entire files with hundreds of lines of code, enabling the discovery of novel algorithms for matrix multiplication, solutions to open mathematical problems, kernel optimization, and more. Although these methods often outperform existing human-designed baselines within their domains, more ambitious efforts, such as AI Scientist \cite{lu2024aiscientistfullyautomated,yamada2025aiscientistv2workshoplevelautomated}, aim for fully autonomous end-to-end scientific discovery with minimal human intervention. While such systems demonstrate impressive capabilities, they still fall short of matching human-level performance (e.g., producing research papers that are accepted at top-tier conferences). To support progress in this direction, MLGym \cite{nathanimlgym} introduces a suite of benchmark environments specifically designed for training and evaluating models on scientific discovery tasks, signaling a promising future for this emerging field.

\begin{figure*}[!ht]
    \centering
    \includegraphics[width=\linewidth, keepaspectratio]{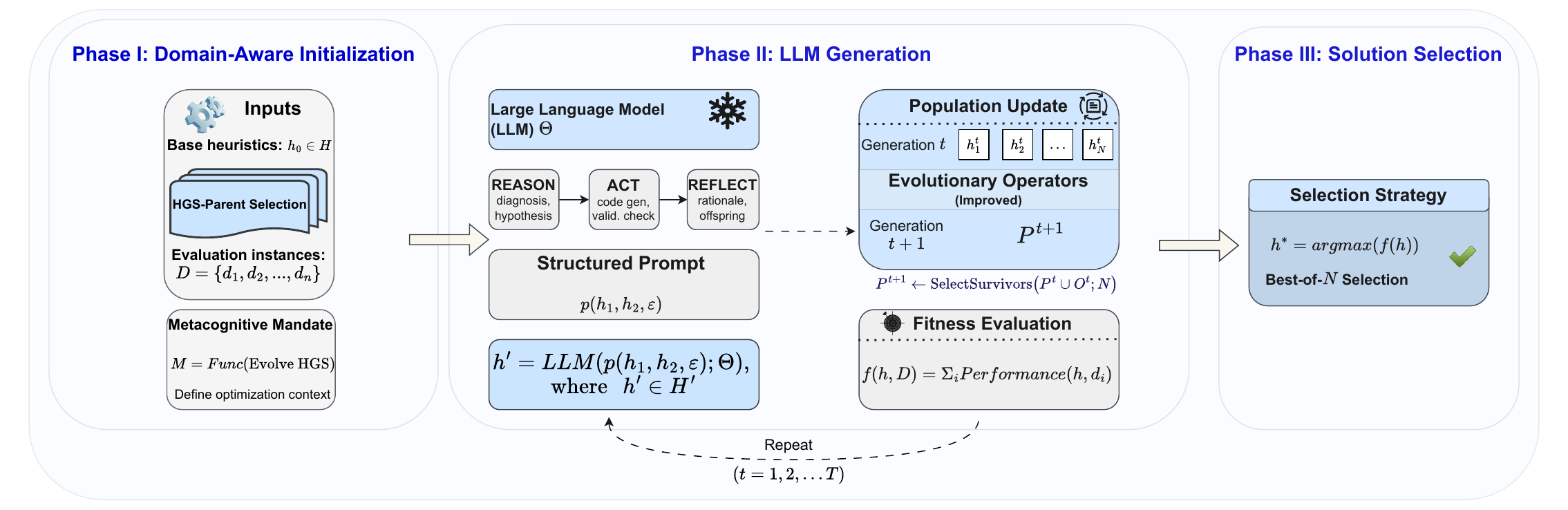}
    \caption{An overview of MEP, which outlines a process for evolving high-performing heuristics. The process continues until convergence criteria or a maximum number of generations is reached, ultimately returning the best-performing heuristic.}
    \Description{Diagram showing the process of evolving heuristics in MEP, including steps for convergence and selection of the best heuristic.}
    \label{fig:heuristic-method}
\end{figure*}


\section{Preliminaries}
\label{sec:prelims}

This section presents the necessary background on Vehicle Routing Problems and the Hybrid Genetic Search algorithm.

\subsection{Vehicle Routing Problem}

The VRP is a class of combinatorial optimization problems focused on finding optimal sets of routes for a fleet of vehicles to serve a set of customers. A VRP is typically defined on a graph $G = (V, A)$, where $V$ is a set of nodes and $A$ is a set of arcs. The node set $V$ is partitioned into a set of depots and a set of customers. Each arc $(i,j) \in A$ is associated with a travel cost $c_{i,j}$. The objective is to determine a set of vehicle routes, starting and ending at a depot, that visit all customers while minimizing a specific objective function, usually the total travel cost. To systematically handle the diverse landscape of VRP variants, we adopt the composable attribute framework used in \cite{berto2023rl4co,zhou2024mvmoe,berto2025routefinder}. 
This allows any variant to be defined by a combination of fundamental properties, facilitating structured analysis and evaluation. These attributes are categorized as follows:

\begin{itemize}[leftmargin=*]
    \item \textbf{Node-specific attributes}, which define local requirements for each customer or depot. The primary examples are:
    \begin{itemize}
        \item \textbf{Demand}: The quantity of goods for pickup or delivery at a node $i$, denoted as $q_i$. This directly influences vehicle load throughout a route.
        \item \textbf{Time Windows (TW)}: The required service interval $[a_i, b_i]$ for a customer, which imposes critical temporal constraints on the route schedule.
        \item \textbf{Service Time}: The duration $s_i$ required to complete service at a node, which consumes time and affects the feasibility of subsequent stops.
    \end{itemize}

    \item \textbf{Global attributes}, which apply to entire routes or the problem as a whole. Key global constraints include:
    \begin{itemize}
        \item \textbf{Capacity (C)}: The maximum load $Q_k$ that a vehicle can carry, limiting the total demand of customers on a single route.
        \item \textbf{Duration Limits (L)}: A maximum allowable travel distance or time for a route, ensuring operational constraints are met.
        \item \textbf{Backhauling (B/MB)}: Rules governing routes with both deliveries (linehauls) and pickups (backhauls), often enforcing precedence constraints.
        \item \textbf{Open Routes (O)}: A structural variant where vehicles are not required to return to the depot after their final delivery.
        \item \textbf{Multi-Depot (MD)}: Scenarios involving multiple starting and ending points for vehicles, adding a layer of assignment complexity.
    \end{itemize}
\end{itemize}

This unified taxonomy provides a clear and systematic method for representing complex VRPs, enabling the evaluation of algorithmic components across a wide spectrum of problem types.


\subsection{VRP Variants Considered}
In this work, we evaluate our approach on a diverse set of VRPs to test the generalization of the discovered heuristics. We consider the following 7 VRP variants:

\begin{itemize}
    \item \textbf{Traveling Salesman Problem (TSP)}: The TSP represents the simplest case of the VRP family, involving a single vehicle (or salesman) that must visit every customer exactly once and return to the starting depot. Despite its conceptual simplicity, the TSP is NP-hard and serves as a building block for more complex routing problems. The objective is to minimize the total travel distance or cost of the tour.
    
    \item \textbf{Capacitated VRP (CVRP)}: The CVRP extends the basic VRP by introducing vehicle capacity constraints. Each vehicle $k$ has a finite capacity $Q_k$, and each customer $i$ has a demand $q_i$ that must be satisfied. The constraint ensures that the total demand of all customers served by a single route cannot exceed the vehicle's capacity: $\sum_{i \in R_k} q_i \leq Q_k$, where $R_k$ represents the set of customers served by vehicle $k$. This variant is fundamental in logistics applications where vehicles have physical limitations.

    \item \textbf{Generalized Vehicle Routing Problem (GVRP)}: In this variant, graph nodes are grouped into mutually exclusive clusters, and serving any single node within a cluster satisfies the service requirement for the entire cluster. This creates a combinatorial challenge where the algorithm must select both which clusters to visit and which specific node within each cluster to serve. The generalized structure is applicable to scenarios where customers have multiple service locations or where service points can substitute for each other.

    \item \textbf{Multi-Depot VRP with Time Windows (MDVRPTW)}: This variant combines the complexity of multiple depot locations with time window constraints. Each vehicle is based at one of several depots and must serve customers within their specified time windows $[a_i, b_i]$. The challenge lies in optimally assigning customers to depots and vehicles while respecting both spatial proximity and temporal feasibility. Each depot typically has its own fleet of vehicles, and routes must start and end at the same depot. This variant is particularly relevant for logistics companies with multiple distribution centers serving geographically dispersed customers with delivery time requirements.
    
    \item \textbf{Prize-Collecting VRP with Time Windows (PCVRPTW)}: Unlike traditional VRP variants where all customers must be visited, the PCVRPTW allows for selective customer service. Each customer $i$ has an associated prize $\pi_i$ collected when visited, and the objective is to maximize the total collected prize minus the total travel cost, subject to a total cost budget $B$. Customers have time windows, and the challenge is to identify the most profitable subset of customers to serve within the available resources.

    \item \textbf{VRP with Backhauls (VRPB)}: The VRPB addresses scenarios where customers are divided into two categories: linehaul customers (requiring delivery from the depot) and backhaul customers (requiring pickup to the depot). A critical constraint is that all linehaul customers must be served before any backhaul customers on the same route. This precedence constraint reflects practical considerations in logistics, where vehicles must deliver goods before they have space to collect returns or recyclables.
    
    \item \textbf{VRP with Time Windows (VRPTW)}: Each customer $i$ must be visited within a specified time window $[a_i, b_i]$, where $a_i$ is the earliest service time and $b_i$ is the latest service time. If a vehicle arrives before $a_i$, it must wait until the service can begin. Arrival after $b_i$ renders the solution infeasible. This variant is particularly relevant in applications such as package delivery, where customers have preferred delivery times, or in service industries with appointment scheduling.

\end{itemize}

A formal mathematical definition of the base VRP is provided in Appendix \ref{append:vrp_attribute}.

\subsection{Hybrid Genetic Search}

HGS is a powerful metaheuristic framework that combines the global search capabilities of a genetic algorithm with the fine-tuning power of a highly effective local search. The general workflow of HGS is as follows:

\begin{itemize}

\item \emph{Initialization:} A diverse initial population is generated, often using problem-specific constructive heuristics.

\item \emph{Parent Selection:} Two parent solutions ($P_1$,$P_2$) are selected using a fitness-biased mechanism that balances solution quality (elitism) and genetic diversity, typically measured by solution distance metrics.

\item \emph{Crossover:} An offspring solution ($C$) is generated using problem specific recombination operators, such as Selective Route-EXchange (SREX) for VRPs, that effectively preserve meaningful structural characteristics of the parent solutions.

\item \emph{Education (Local Search):} The offspring undergoes intensive improvement via a problem-tailored local search procedure. For VRPs, this typically involves neighborhood moves like \textit{relocate}, \textit{swap}, and \textit{2-opt} operators. This hybridization with local search is fundamental to HGS's performance.

\item \emph{Survivor Selection:} The educated offspring is considered for population inclusion. If population size limits are exceeded, removal prioritizes either low-quality solutions or those exhibiting high similarity to others, maintaining both quality and diversity through a distance-based diversity management mechanism.

\item \emph{Diversification:} When stagnation is detected, the search restarts through partial population reinitialization or injection of new diverse solutions, preventing premature convergence.

\end{itemize}

PyVRP is a state-of-the-art, open-source implementation of HGS specific to VRPs, written in Python with performance-critical components in C++. Its modular architecture enables isolated replacement of core components (e.g., parent selection, local search operators) with custom implementations, creating an ideal experimental platform for heuristic evolution via LLM-generated functions.


\section{Metacognitive Evolutionary Programming}

We propose Metacognitive Evolutionary Programming (MEP), an evolutionary framework inspired by \citet{bai2025mp} that emulates human metacognitive reasoning by guiding the LLM through a structured and strategic design process. As illustrated in Figure \ref{fig:heuristic-method}, MEP enables the LLM to evolve heuristics through a human-like reason–act–reflect cycle, augmented with domain knowledge and guided by a planning strategy that fosters strategic improvement. 

The central thesis of MEP is that engaging the LLM in a structured, hypothesis-driven process makes code evolution more efficient and more likely to yield novel, high-performing heuristics. The framework is composed of two main phases: a one-time \emph{Domain-Aware Initialization}, and an iterative \emph{Reason–Act–Reflect cycle}.

\subsection{Phase 1: Domain-Aware Initialization}

Before the evolutionary search begins, MEP equips the LLM with comprehensive knowledge about the target component. This planning phase provides three critical knowledge types:

\begin{itemize}
    \item \emph{Common Pitfalls ($K_p$):} Known failure modes specific to the component being evolved.
    
    \item \emph{Mitigation Strategies ($K_s$):} Proven techniques to address identified pitfalls.
    
    \item \emph{Problem-Specific Traps ($K_t$):} Domain-specific challenges (e.g., in VRP solving).
\end{itemize}

This knowledge foundation serves as domain expertise that guides the LLM's reasoning throughout the evolutionary process, ensuring that generated heuristics are informed by established algorithmic principles.

\subsection{Phase 2: Reason-Act-Reflect Cycle}

Each generation of a new heuristic in MEP follows a structured three-step cognitive process, called the Reason-Act-Reflect cycle.

\paragraph{Step 1: REASON (Diagnosis and Hypothesis)}
At the start of each evolutionary timestep, two parent heuristics are selected from the current population. Before generating any code, the LLM is prompted to perform a structured analysis of the parents, encouraging deliberate reasoning over random mutation.

\begin{itemize}
    \item \emph{Diagnosis:} The LLM begins by analyzing the provided parent heuristics, i.e., \texttt{code\textsubscript{1}}, \texttt{code\textsubscript{2}}, along with their performance scores and any available feedback. It is explicitly prompted to diagnose their weaknesses, guided by the principles introduced in \emph{Phase 1}.

    \item \emph{Design Hypothesis:} Building on the identified weaknesses, the LLM must formulate a concise, one-sentence \textit{Design Hypothesis} that proposes a specific improvement for the new heuristic (\texttt{code\textsubscript{3}}). This step is crucial, as it ensures the LLM explicitly reasons about how to address the limitations of its predecessors, rather than simply combining elements of existing code.

\end{itemize}

\paragraph{Step 2: ACT (Code Implementation)} Once the hypothesis is formulated, the LLM is tasked with implementing it. This step is governed by strict guardrails including a fixed function signature, approved libraries, and a required output format, to ensure the generated code is syntactically valid and fully compatible with the PyVRP evaluation sandbox.

\paragraph{Step 3: REFLECT (Rationale and Self-Critique)} After generating the code, the LLM engages in a final metacognitive step: self-reflection. In this phase, it critically evaluates its prior reasoning, hypothesis formulation, and implementation choices. This self-assessment is recorded and carried forward to inform the generation of future offspring, enabling improvement over subsequent evolutionary cycles.

\begin{figure}[!ht]
\centering
\fbox{
\begin{minipage}{0.465\textwidth}
\footnotesize 
\textbf{Role: AI Optimization Researcher (Vehicle Routing Problem)}
\hrule
\vspace{2mm}
\textbf{\textsc{1) Planning Phase}}
\begin{itemize}
    \setlength\itemsep{0em}
    \item \textbf{Pitfalls ($K_p$):} List potential failures in VRP parent selection...
    \item \textbf{Strategies ($K_s$):} Propose countermeasures for each pitfall...
    \item \textbf{Hidden Traps ($K_t$):} Identify misleading problem instance features...
\end{itemize}

\textbf{Overall Objective:} Write a \texttt{novel} Python function \texttt{select\_parents} for a Hybrid Genetic Search algorithm, guided by the planning insights.

\vspace{2mm}
\textbf{Context:} Background on parent selection in genetic algorithms for VRP, including standard approaches (e.g., elitist, tournament). ...

\vspace{2mm}
\textbf{Common Existing Approaches:} Standard parent selection methods in genetic algorithms include elitist selection (top-ranked individuals), tournament selection (best from random subset), fitness-proportionate selection, etc. ...

\vspace{2mm}
\textbf{\textsc{2) Reasoning Phase}}
\begin{itemize}
    \setlength\itemsep{0em}
    \item \textbf{Analyze Patterns:} Assess exploration/exploitation bias in provided selectors.
    \item \textbf{Assess Shortcomings:} Evaluate selectors against the $K_p$/$K_t$ list.
    \item \textbf{Design Sketch:} Outline the proposed new selector, \texttt{sel3}, in prose.
\end{itemize}
\vspace{2mm}
\textbf{Your Task in This Interaction:}
\begin{enumerate}
    \setlength\itemsep{0em}
    \item Analyze \texttt{sel1} and \texttt{sel2}.
    \item Design and implement \texttt{sel3 = select\_parents(...)}.
    \item \textbf{Implementation Requirements:}
    \begin{itemize}
        \setlength\itemsep{0em}
        \item \textbf{Inputs:} \texttt{population}, \texttt{rng}, \texttt{cost\_evaluator}, \texttt{k=2}
        \item \textbf{Output:} \texttt{tuple[Solution, Solution]}
        \item \textbf{Behavior:} Balance feasibility, cost, and diversity.
    \end{itemize}
\end{enumerate}
\vspace{2mm}
\textbf{Reference Python Classes:}
\begin{quote}
\ttfamily
\begin{tabular}{@{}l@{}}
class CostEvaluator: ... \\
class Solution: ... \\
class ProblemData: ... \\
\# ... (other PyVRP class definitions)
\end{tabular}
\end{quote}

\textbf{\textsc{3) Reflection Phase}}
\begin{enumerate}
    \setlength\itemsep{0em}
    \item How does the final code address the initial pitfalls ($K_p$)?
    \item Did new challenges emerge during coding?
    \item Propose two concrete revisions for further improvement.
\end{enumerate}
\vspace{2mm}
\textbf{Response Format:} A single Python code block containing only the required function and imports.
\vspace{2mm}
\hrule
\vspace{2mm}
\textbf{Dynamic Inputs for a Given Task:}
\begin{itemize}
    \setlength\itemsep{0em}
    \item \textbf{Selector 1:} \texttt{\$code1}
    \item \textbf{Performance:} Cost: \texttt{\$score1}, Feedback: \texttt{\$feedback1}
    \item \textbf{Selector 2:} \texttt{\$code2}
    \item \textbf{Performance:} Cost: \texttt{\$score2}, Feedback: \texttt{\$feedback2}
\end{itemize}
\end{minipage}
}
\caption{The prompt template used to guide the LLM. It consists of distinct instructions for planning, reasoning, and reflection, providing both high-level guidance and detailed technical specifications.}
\label{fig:prompt_template}
\vspace{-2mm}
\end{figure}

\begin{table*}[!ht]
\centering
\caption{Average cost comparison of the original PyVRP components and the best components evolved by MEP on 100 TSP instances. Lower cost values indicate better performance.}
\label{tab:component_fitness}
\renewcommand\arraystretch{1.5}
\begin{tabular}{l S[table-format=7.0] S[table-format=7.0] S[table-format=1.2,table-space-text-post=\%]}
\toprule
\textbf{Component Evolved} & {\textbf{Baseline Cost}} & {\textbf{Best Evolved Cost}} & {\textbf{Improvement}} \\
\midrule
\texttt{select\_parents}   & 7942166 & \textbf{7764815} & \textbf{2.23\,\%} \\
\texttt{select\_survivors}   & 7942166 & \textbf{7886852} & \textbf{0.69\,\%} \\
\texttt{update\_penalties} & 7942166 & \textbf{7932973} & \textbf{0.12\,\%} \\
\bottomrule
\end{tabular}
\end{table*}

The Reason-Act-Reflect cycle enforces a principled generation process, ensuring that each new heuristic emerges from structured reasoning rather than arbitrary mutation. Figure \ref{fig:heuristic-method} outlines the process for evolving high-performing heuristics using the proposed method. A high-level prompt structure is presented in Figure \ref{fig:prompt_template}, breaking-down the LLM interaction into three distinct phase-aligned segments. \emph{Planning} provides high-level strategic context, explicitly injecting domain knowledge to ground the LLM's initial reasoning in established principles. \emph{Reasoning} directs the LLM to perform a structured diagnosis of parent heuristics' weaknesses and mandates the formulation of a concise, testable \emph{Design Hypothesis} before any code generation, ensuring deliberate improvement intent. \emph{Reflection} requires the LLM to self-critique its generated code and the preceding reasoning/hypothesis within the code documentation, embedding metacognitive assessment directly into the output. Each segment combines high-level instructions (e.g., ``Diagnose the weaknesses...'') with detailed technical specifications (e.g., fixed function signatures, approved libraries, required output format) to enforce both strategic coherence and syntactic correctness. The full prompt is provided in Appendix \ref{append:full_llm_prompt}. 


\section{MEP Case Study on HGS}

We apply MEP to evolve three critical components within the HGS algorithm. This includes \texttt{select\_parents} for fitness-diversity balanced selection, 
\texttt{select\_survivors} for population management, and \texttt{update\_penalties} for constraint handling of infeasible solutions. These operators directly govern the exploration-exploitation trade-off of the genetic algorithm. In this context, exploitation refers to selecting high-quality (low-cost) parents to refine the best-known solutions, while exploration involves selecting diverse parents to introduce new genetic material and prevent premature convergence to local optima. A better heuristic must balance these two competing objectives. To this end, we use the \textbf{PyVRP} library as our experimental sandbox. Thanks to its modular design, we can isolate and replace core HGS components (such as the parent selection operator) with LLM-generated functions. This approach creates \our, a collection of better-performing heuristics.

\paragraph{Evolutionary Process:} It is outlined as follows:

\begin{itemize}
    \item \textbf{Base Population:} The base population consists of default implementations of selected modules from PyVRP.
    
    \item \textbf{Evaluation:} Each heuristic in the population is evaluated by running HGS on a set of VRP instances, with its fitness score defined as the negative average cost of the solution achieved.
    
    \item \textbf{Generation:} In each generation, two parent heuristics and their performance scores are selected. These, together with domain knowledge, are provided to the MEP prompt to produce a new offspring heuristic.
    
    \item \textbf{Population Update:} The new heuristic is evaluated and considered for inclusion in the population.
    
    \item \textbf{Termination:} The process runs for a fixed number of generations (N), and the best heuristic discovered is reported.
\end{itemize}

The evolutionary process runs for 10 generations. In each generation, 10 offspring heuristics are produced, and the top 5 performers, i.e., selected from both the new offspring and the existing population, are retained for the next generation. To ensure the robustness of our findings, the entire evolutionary process for each component was conducted five times using different random seeds, and the best-performing heuristic from these runs is reported in our results. The LLM interactions were performed using the GPT-4.1 model. We set the temperature parameter to 1.0 to encourage creative and diverse heuristic designs, while all other parameters, such as top-p and max\_tokens, were left at their default values.

\paragraph{Benchmark Suite:} We evaluate our approach on two datasets. First, we use the publicly available TSP100 dataset \cite{yao2024rethinking}, a well-established benchmark in combinatorial optimization, to assess LLM-generated heuristics during the evolution process. Second, we evaluate the discovered HGS variants on a comprehensive benchmark suite spanning six VRP variants.
Each variant, sourced from the PyVRP library\footnotemark[1], includes a dedicated folder containing 60 instances, ensuring statistically robust evaluation. The benchmark spans a range of problem complexities, i.e., from the standard Capacitated VRP (CVRP) to more challenging formulations like the Multi-Depot VRP with Time Windows (MDVRPTW). 
The diversity of these variants, each with distinct constraint structures and objectives, provides a comprehensive testing ground for evaluating the robustness and generalizability of evolved heuristics. \footnotetext[1]{https://github.com/PyVRP/Instances/}


\section{Experimental Result and Analysis}

Our primary goal is to demonstrate that MEP can discover better heuristic components for a SOTA VRP solver. Before presenting the results, it is crucial to distinguish between two different time costs: the one-time design cost of the evolutionary process and the recurring execution cost of the final heuristic. The performance tables in our analysis report the execution cost, which is the wall-clock time required for the final, evolved heuristic to solve a given VRP instance. This metric allows for a direct comparison against the baseline solver's runtime. In contrast, the design cost is the upfront, one-time computational investment required to discover the heuristic. This involves the entire MEP process, including all LLM API calls and the evaluation of candidate solutions over multiple generations. We consider this design cost analogous to the manual effort a human researcher would invest over weeks or months to develop a novel algorithm. For this work, the total design cost for discovering each final component was approximately 3–4 hours on a single NVIDIA H100 GPU. This involved approximately 100 LLM API calls per component at a total cost of ~\$15–20 USD, substantially lower than the weeks of manual effort typically required. We posit that this one-time investment is a practical trade-off for producing novel heuristics that significantly improve solution quality and reduce subsequent execution costs.
We use GPT-4.1 in the evolutionary search, as mentioned earlier, chosen for its strong performance and cost-effectiveness. The following standard metrics are used to evaluate solution quality:

\begin{itemize}
    \item \textbf{Cost/Obj.:} The total cost of a solution, typically representing the sum of travel distances or travel times across all routes. Lower values indicate better solutions.
    
    \item \textbf{Improvement:} The relative percentage by which the method's objective value is better than the best-known solution (BKS). It is computed as:
\begin{equation}
    \text{Improvement} = \frac{\text{Obj}_{\text{BKS}} - \text{Obj}_{\text{method}}}{\text{Obj}_{\text{BKS}}} \times 100\%
\end{equation}
\end{itemize}

\begin{table*}[!ht]
\centering
\caption{Performance comparison of the baseline PyVRP solver versus the \our equipped with the MEP-evolved operators across various VRP variants (average values are presented). Lower cost and shorter times indicate better performance.}
\renewcommand\arraystretch{1.0}
\label{tab:variant_generalization}
\sisetup{round-mode=places, round-precision=2, table-format=7.2}
\begin{tabular}{l S S S[table-format=1.2,table-space-text-post=\%] S[table-format=2.2] S[table-format=2.2]}
\toprule
\textbf{Variant} & {\textbf{Baseline Cost}} & {\textbf{MEP-Evolved Cost}} & {\textbf{Improvement}} & {\textbf{Baseline Time (s)}} & {\textbf{MEP-Evolved Time (s)}} \\
\midrule
CVRP      & 64450.03    & \textbf{64236.30}    & 0.33\,\% & 9.55 & \textbf{9.11}  \\
GVRP      & 7686993.64  & \textbf{7532007.09}  & 2.01\,\% & \textbf{0.87}  & 1.89           \\
MDVRPTW   & 8955.11     & \textbf{8825.46}     & 1.45\,\% & 12.43 & \textbf{9.74}  \\
PCVRPTW   & 22975.85    & \textbf{22406.42}    & 2.48\,\% & 7.92 & \textbf{4.19}  \\
VRPB      & 83696.21    & \textbf{83146.06}    & 0.66\,\% & 14.07 & \textbf{13.85} \\
VRPTW     & 33424.48    & \textbf{32521.75}    & 2.70\,\% & 17.03 & \textbf{14.66} \\
\bottomrule
\end{tabular}
\end{table*}

A positive value indicates that the method outperforms the BKS, while a negative value means it performs worse. 

\subsection{Hardware Settings}
All experiments are conducted on a high-performance computing cluster with the following specifications:

\begin{itemize}
  \item \textbf{GPU:} NVIDIA H100 GPU with dedicated allocation
  \item \textbf{CPU:} 32-core processor allocation per job
  \item \textbf{Memory:} 120\,GB RAM per compute node
  \item \textbf{Cluster Configuration:} Single-node allocation with GPU acceleration
\end{itemize}

\subsection{Evolving Effective HGS Components}
\begin{figure}[h!]
    \centering
    \includegraphics[width=0.48\textwidth]{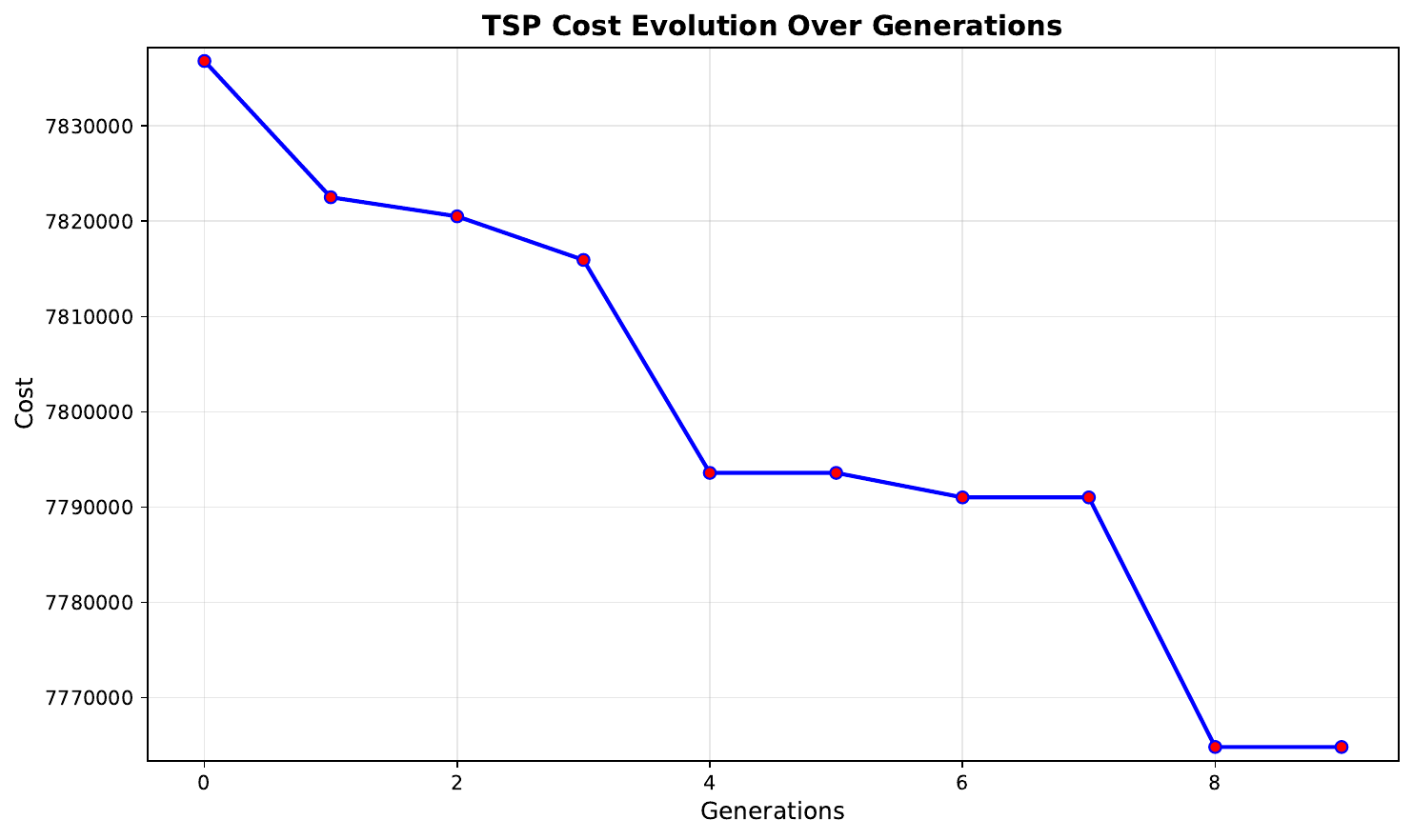}
    \caption{Evolution of average cost values over 10 generations for TSP optimization using MEP for parent selection, with averages computed across 100 instances.
    }
    \label{fig:fitness_evolution}
\end{figure}

Table~\ref{tab:component_fitness} summarizes the average cost of the best-evolved heuristic for each target component compared to the PyVRP baseline on the TSP100 benchmark. In all cases, MEP successfully discovered an operator with a better cost. The most significant improvements were observed in the \texttt{select\_parents} and \texttt{select\_survivors} components, which saw a 2.23\% and 0.69\% improvement, respectively. This highlights MEP's ability to innovate on complex, high-impact components of the metaheuristic. The cost-generation curve illustrating the iterative improvement of the evolutionary process is shown in Figure~\ref{fig:fitness_evolution}.

A key outcome of the MEP process is the emergence of sophisticated and novel heuristics that go beyond simple variations of existing strategies. For example, the top-performing \texttt{select\_parents} operator, described by the LLM as a ``Hybrid Adaptive Stratified Diversity-Pressure Selection'', divides the population into cost-based strata and applies tailored diversity pressures to each subgroup. 
The complete evolved code is provided in Appendix \ref{append:parent_selection_code}. These results highlight that MEP’s hypothesis-driven framework enables the discovery of complex, well-motivated algorithmic innovations, far beyond simple parameter tuning or superficial modifications.

To evaluate the synergistic potential of these evolved heuristics, we created an integrated HGS solver that combines the best-performing LLM-generated modules for parent selection, survivor selection, and penalty adjustment. This integrated solver represents a more holistic test of MEP's capabilities, assessing whether components evolved in isolation can cooperate effectively. As we will demonstrate in the following section, this integrated approach not only matches but often exceeds the performance of individually evolved modules, achieving consistent cost reductions and significant runtime improvements on complex VRP variants. This finding validates that MEP can produce a suite of compatible, high-performing heuristics that collectively advance the state-of-the-art.


\subsection{Integrated Solver Performance and Generalization}


To assess the robustness and overall performance of our approach, we evaluated the fully integrated HGS solver, i.e., equipped with the best evolved heuristics for parent selection, survivor selection, and penalty updates against the PyVRP baseline on a diverse set of VRP variants. The results, shown in Table~\ref{tab:variant_generalization}, demonstrate a clear and consistent advantage for the MEP-evolved solver.

The integrated solver achieves notable cost reductions across all variants, with particularly strong improvements on complex instances like VRPTW (2.70\%), PCVRPTW (2.48\% improvement) and GVRP (2.01\% improvement). Furthermore, the evolved heuristics lead to significant computational efficiencies, reducing runtime by over 45\% on PCVRPTW and 10\% on MDVRPTW. While some variants show a modest increase in runtime, the superior solution quality underscores the effectiveness of the discovered strategies. This strong generalization performance confirms that MEP is capable of producing a synergistic set of heuristics that are not over-fitted to a single problem type and can enhance a state-of-the-art solver across a wide range of operational scenarios.

\begin{table}[h!]
\centering
\caption{Ablation study results for the \texttt{select\_parents} operator, showing percentage improvement over the baseline. Average cost values are presented in the table.}
\label{tab:ablation_study}
\sisetup{
  round-mode=places, 
  round-precision=2,
  table-format=-1.2,
  table-space-text-post=\%
}
\renewcommand\arraystretch{1.5}
\resizebox{0.48\textwidth}{!}{ 
\begin{tabular}{l S[table-format=7.2] S S S}
\toprule
\textbf{Variant} & {\textbf{Baseline Cost}} & {\textbf{Full MEP Impr.}} & {\textbf{MEP-noInit Impr.}} & {\textbf{Reactive Evol. Impr.}} \\
\midrule
CVRP      & 64450.03    & \textbf{0.33\,\%} & 0.28\,\% & -0.20\,\% \\
GVRP      & 7686993.64  & \textbf{1.82\,\%} & 0.98\,\% & 1.30\,\% \\
MDVRPTW   & 8955.11     & \textbf{1.49\,\%} & 0.90\,\% & -2.05\,\% \\
PCVRPTW   & 22975.85    & \textbf{2.43\,\%} & 2.36\,\% & 1.46\,\% \\
VRPB      & 83696.21    & \textbf{0.57\,\%} & 0.53\,\% & -0.39\,\% \\
VRPTW     & 33424.48    & \textbf{2.25\,\%} & 2.24\,\% & 0.10\,\% \\
\bottomrule
\end{tabular}}
\end{table}

\begin{table}[!ht]
\centering
\caption{Generalization performance of individual versus combined MEP-evolved operators. The table shows the percentage improvement over the HGS baseline, with the final column (\our) representing a solver that integrates all individually evolved components.}
\label{tab:other_generalization}
\sisetup{
  round-mode=places, 
  round-precision=2,
  table-format=-1.2,
  table-space-text-post=\%
}
\renewcommand\arraystretch{1.5}
\resizebox{0.48\textwidth}{!}{ 
\begin{tabular}{l S S S S}
\toprule
\textbf{Variant} & {\textbf{select\_parents}} & {\textbf{select\_survivors}} & {\textbf{update\_penalties}} & {\textbf{\our}}\\
\midrule
CVRP      & \textbf{0.33\,\%} & 0.05\,\% & 0.01\,\% & \textbf{0.33\,\%} \\
GVRP      & 1.82\,\% & 0.08\,\% & 0.13\,\% & \textbf{2.01\,\%} \\
MDVRPTW   & \textbf{1.49\,\%} & -0.00\,\% & 0.16\,\% & 1.45\,\% \\
PCVRPTW   & 2.43\,\% & 0.15\,\% & 0.06\,\% & \textbf{2.48\,\%} \\
VRPB      & 0.57\,\% & 0.04\,\% & 0.05\,\% & \textbf{0.66\,\%} \\
VRPTW     & 2.25\,\% & 0.29\,\% & 0.35\,\% & \textbf{2.70\,\%} \\
\bottomrule
\end{tabular}}
\end{table}

\subsection{Generalization Across Evolved Components}

We also evaluated the generalization capabilities of the other best-evolved. Two of these experimental results are presented in Table \ref{tab:other_generalization}, which also demonstrate positive, albeit more modest, improvements over the baseline across various VRP instances. For example, the evolved `select\_survivors` operator improves VRPTW performance by 0.29\%, while the evolved `update\_penalties` operator improves it by 0.35\%. This indicates that the MEP framework is capable of discovering generally beneficial heuristics for multiple components of the HGS algorithm, even if the primary gains are concentrated in the highest-impact operators, such as parent selection.

Notably, none of the evolved HGS components underperformed their baseline counterparts on any of the six VRP variants, despite not being explicitly evaluated on these during the evolution process. This highlights the robustness and generalizability of the discovered algorithmic strategies.

\subsection{Ablation Studies}

To assess the individual contributions of key components in our framework, we performed a series of ablation studies focused on the select\_parents module, chosen for its strong generalization across tasks. These experiments aim to isolate the effects of the Domain-Aware Initialization and the structured Reason–Act–Reflect cycle. Our ablation compares full MEP framework against two variants:
\begin{itemize}
    \item MEP-noInit: This version removes the Domain-Aware Initialization from the prompt. The LLM still performs the Reason-Act-Reflect cycle, but its reasoning is purely local, based only on the two parent heuristics provided in each generation. 
    \item Reactive Evolution: This variant removes the structured reasoning components. The prompt is simplified, providing the parent heuristics and their scores and asking the LLM to generate an improved version. This setup mimics prior reactive evolution methods such as EoH \cite{liu2024evolution} and ReEvo \cite{ye2024reevo}.
\end{itemize}

The results in Table~\ref{tab:ablation_study} reveal a clear performance hierarchy. The full MEP framework consistently achieves the best results. The MEP-noInit variant performs reasonably well but is worse than the full MEP, highlighting the benefit of the strategic guidance provided by the Domain-Aware Initialization. For example, on GVRP, the improvement drops from 1.82\% with the full MEP to 0.98\% without the Initialization. The Reactive Evolution variant performs the worst, often yielding only marginal gains or even performance degradation (e.g., -0.20\% on CVRP), underscoring the importance of the structured reason-act-reflect cycle.


\section{Conclusion}

In this work, we introduced Metacognitive Evolutionary Programming (MEP), a new paradigm for automated algorithm design that enhances LLM-driven evolution with strategic planning and structured reasoning for the Vehicle Routing Problem (VRP). By grounding the LLM in domain knowledge and guiding it through a structured Reason–Act–Reflect cycle, we elevate it from a basic code mutator to a strategic agent capable of deliberate heuristic discovery. Our application of MEP to evolve core components of the state-of-the-art HGS solver demonstrates a promising path toward discovering heuristics that are not only high-performing but also novel and strategically sound. Looking ahead, this work opens up several avenues for future research. 
Our work demonstrates that MEP can successfully enhance individual components of a state-of-the-art solver. More importantly, we have shown that these independently evolved heuristics can be successfully integrated, creating a holistically improved solver that generalizes across multiple challenging VRP variants. Future work could explore the joint, simultaneous evolution of all components to uncover even deeper synergistic interactions, though this remains a computationally intensive challenge.
While this presents significant challenges with the current LLM context length and computational costs, we believe it is crucial for uncovering synergistic interactions between components. 
Our work demonstrates that a manageable one-time design cost can yield heuristics with lasting performance benefits, providing a strong incentive to tackle these future challenges. Furthermore, to rigorously test the generalizability of our discovered heuristics, we plan to evaluate their performance across the full set of 48 VRP variants covered in the original PyVRP. 
Overall, this work lays a foundation for more autonomous and intelligent algorithmic systems capable of scientific discovery, transcending the constraints imposed by manually designed algorithms. The code is available at \url{https://github.com/ra-MANUJ-an/pyvrp-code}.



\begin{acks}
This research is supported by the National Research Foundation, Singapore under the AI Singapore Programme (AISG Award No: AISG3-RPGV-2025-017), and the Singapore Ministry of Education (MOE) Academic Research Fund (AcRF) Tier 1 grant.
\end{acks}


\newpage
\bibliographystyle{ACM-Reference-Format} 
\balance
\bibliography{sample}


\newpage
\appendix


\section{VRP Attribute Taxonomy}
\label{append:vrp_attribute}

To handle the vast landscape of VRP variants in a structured manner, we describe VRP instances via composable attributes. This allows us to represent any variant as a combination of fundamental properties, enabling systematic generation and evaluation of evolved components across a multitude of problem types. The attributes are categorized into node-specific and global properties, each with precise mathematical formulations. Let $G = (V, A)$ be a directed graph where $V = \{0, 1, 2, \ldots, n\}$ is the set of nodes with node 0 representing the depot and $\{1, 2, \ldots, n\}$ representing customers, and $A = \{(i,j) : i, j \in V, i \neq j\}$ is the set of arcs. Let $m$ denote the number of available vehicles.

\paragraph{Node attributes:} Properties specific to each customer or depot node $i \in V$. These attributes define the local requirements and characteristics that must be satisfied during service.

\begin{itemize}
\item \textbf{Demand:} The demand attribute represents the quantity of goods to be delivered to or picked up from each customer. Let $q_i \in \mathbb{R}$ denote the demand at node $i$, where:
\begin{equation}
q_i = \begin{cases}
> 0 & \text{if customer } i \text{ requires delivery (linehaul)} \\
< 0 & \text{if customer } i \text{ requires pickup (backhaul)} \\
= 0 & \text{if node } i \text{ is the depot}
\end{cases}
\end{equation}

\item \textbf{Time Window (TW):} Each customer $i$ has a designated time window $[a_i, b_i]$ within which service must commence. Let $t_i$ denote the service start time at customer $i$. The time window constraints are:
\begin{align}
a_i &\leq t_i \leq b_i, \quad \forall i \in V \setminus \{0\} \label{eq:time_window}
\end{align}
\begin{align}
t_i + s_i + \tau_{ij} &\leq t_j + M(1 - x_{ijk}), \\
&\quad \forall i,j \in V, k \in \{1,\ldots,m\} \label{eq:time_propagation}
\end{align}
where $s_i \geq 0$ is the service time at customer $i$, $\tau_{ij} \geq 0$ is the travel time from $i$ to $j$, $M$ is a sufficiently large constant, and $x_{ijk} \in \{0,1\}$ indicates whether vehicle $k$ traverses arc $(i,j)$.

The depot time window $[a_0, b_0]$ constrains the total route duration:
\begin{equation}
a_0 \leq t_0^{\text{start}} \leq t_0^{\text{end}} \leq b_0
\end{equation}
where $t_0^{\text{start}}$ and $t_0^{\text{end}}$ represent the departure and return times at the depot, respectively.

\item \textbf{Service Time:} Each customer $i$ requires a non-negative service time $s_i \geq 0$ during which the vehicle is occupied. This affects the temporal feasibility of routes and is incorporated into time window constraints through equation \eqref{eq:time_propagation}.

The total service time for vehicle $k$ is:
\begin{equation}
S_k = \sum_{i \in V} s_i y_{ik}, \quad \forall k \in \{1,\ldots,m\}
\end{equation}
where $y_{ik} \in \{0,1\}$ indicates whether customer $i$ is served by vehicle $k$.
\end{itemize}

\paragraph{Global attributes:} Constraints or properties that apply to entire routes or the complete problem instance, affecting the optimization objective and feasibility conditions.

\begin{itemize}
\item \textbf{Capacity (C):} Each vehicle $k$ has a finite capacity $Q_k$. The capacity constraint ensures that the total demand of all customers served by a single route cannot exceed the vehicle's capacity:
\begin{equation}
\sum_{i \in V} q_i y_{ik} \leq Q_k, \quad \forall k \in \{1, \ldots, m\}
\end{equation}

\item \textbf{Duration Limit (L):} This constraint imposes a maximum allowable cost or duration $D_k$ for each vehicle route $k$. The constraint is formulated as:
\begin{align}
\sum_{(i,j) \in A} c_{ij} x_{ijk} + \sum_{i \in V} s_i y_{ik} &\leq D_k, \\
&\quad \forall k \in \{1,\ldots,m\} \label{eq:duration_limit}
\end{align}
where $c_{ij}$ represents the travel cost between nodes $i$ and $j$.

For homogeneous duration limits across all vehicles, $D_k = D$ for all $k$.

\item \textbf{Backhaul (B):} In backhaul problems, customers are partitioned into linehaul customers $L = \{i \in V : q_i > 0\}$ and backhaul customers $B = \{i \in V : q_i < 0\}$. The precedence constraint ensures all linehauls are served before backhauls:
\begin{align}
\sum_{j \in B} \sum_{(i,j) \in A} x_{ijk} &= 0, \\
&\quad \forall i \in L, k \in \{1,\ldots,m\} \label{eq:backhaul_precedence}
\end{align}

Alternatively, using auxiliary variables $u_{ik} \in \mathbb{R}^+$ representing the position of customer $i$ in vehicle $k$'s route:
\begin{align}
u_{ik} + 1 &\leq u_{jk} + n(1 - x_{ijk}), \\
&\quad \forall i \in L, j \in B, k \in \{1,\ldots,m\} \label{eq:backhaul_position}
\end{align}
where $n = |V|$ is the total number of nodes.

\item \textbf{Open Route (O):} In open route problems, vehicles are not required to return to the depot after serving their last customer. This modifies the route structure by eliminating mandatory return arcs. The flow conservation constraints become:
\begin{align}
\sum_{j \in V \setminus \{0\}} x_{0jk} &= z_k, \quad \forall k \in \{1,\ldots,m\} \label{eq:depot_departure}
\end{align}
\begin{align}
\sum_{j \in V} x_{jik} - \sum_{j \in V} x_{ijk} &= 0, \\
&\quad \forall i \in V \setminus \{0\}, k \in \{1,\ldots,m\} \label{eq:flow_conservation_open}
\end{align}
\begin{align}
\sum_{j \in V \setminus \{0\}} x_{j0k} &\leq z_k, \quad \forall k \in \{1,\ldots,m\} \label{eq:optional_return}
\end{align}
where $z_k \in \{0,1\}$ indicates whether vehicle $k$ is used.

\item \textbf{Multi-Depot (MD):} Let $D = \{d_1, d_2, \ldots, d_p\} \subset V$ denote the set of depots. Each vehicle $k$ is assigned to exactly one depot $\text{depot}(k) \in D$:
\begin{equation}
\sum_{d \in D} w_{kd} = 1, \quad \forall k \in \{1,\ldots,m\}
\end{equation}
where $w_{kd} \in \{0,1\}$ indicates whether vehicle $k$ is assigned to depot $d$.

The route must start and end at the assigned depot:
\begin{align}
\sum_{j \in V \setminus D} x_{djk} &= w_{kd}, \quad \forall d \in D, k \in \{1,\ldots,m\}
\end{align}
\begin{align}
\sum_{i \in V \setminus D} x_{idk} &= w_{kd}, \\
&\quad \forall d \in D, k \in \{1,\ldots,m\} \text{ (if not open routes)}
\end{align}

Each customer must be served by exactly one vehicle:
\begin{equation}
\sum_{k=1}^{m} y_{ik} = 1, \quad \forall i \in V \setminus D
\end{equation}

\item \textbf{Mixed Backhaul (MB):} Mixed backhaul relaxes the strict precedence constraint of standard backhaul problems, allowing pickups and deliveries to be intermixed on the same route. Let $\ell_{ik} \geq 0$ represent the load of vehicle $k$ after serving customer $i$. The constraints are:
\begin{align}
0 &\leq \ell_{ik} \leq Q_k, \quad \forall i \in V, k \in \{1,\ldots,m\} \label{eq:load_bounds}
\end{align}
\begin{align}
\ell_{jk} &= \ell_{ik} + q_j - M(1 - x_{ijk}), \\
&\quad \forall (i,j) \in A, k \in \{1,\ldots,m\} \label{eq:load_propagation}
\end{align}
\begin{align}
\ell_{0k} &= 0, \quad \forall k \in \{1,\ldots,m\} \label{eq:depot_load}
\end{align}
\end{itemize}

Using this unified format, any VRP variant can be systematically represented as:
\begin{equation}
\text{VRP Variant} = \text{Base VRP} \cup \mathcal{A}_{\text{node}} \cup \mathcal{A}_{\text{global}}
\end{equation}
where $\mathcal{A}_{\text{node}} \subseteq \{\text{TW, Service Time}\}$ and $\mathcal{A}_{\text{global}} \subseteq \{\text{C, L, B, O, MD, MB}\}$.

\section{Full LLM Prompt}
\label{append:full_llm_prompt}

The complete prompt template used in our experiments for parents selection is provided below. This template dynamically incorporates the two existing parent selectors (\texttt{\$code1} and \texttt{\$code2}) along with their performance metrics to guide the generation of improved solutions. We'll be providing more prompt templates alongside the release of the code. The prompt template for our experiments is structured into the following sequential sections:

\begin{enumerate}[noitemsep, nolistsep]  
  \item \textbf{Role and Planning}
  \item \textbf{Context and Existing Approaches}
  \item \textbf{Reasoning Phase and Task Description}
  \item \textbf{Implementation Instructions}
  \item \textbf{Implementation Requirements}
  \item \textbf{Reference Classes}
  \item \textbf{Reflection Phase and Response Format}
  \item \textbf{Existing Parent Selectors}
\end{enumerate}

\begin{tcolorbox}[colback=gray!5, colframe=gray!50, title=Complete Prompt Template - Role and Planning]
\footnotesize

\textbf{Role: AI Optimization Researcher (Vehicle Routing Problem)}

\vspace{2mm}
\textbf{\textsc{1) Planning Phase}}

\emph{Before you inspect any code, pause and map out your thinking.}
\begin{enumerate}
    \item \textbf{Pitfalls ($K_p$):} List three ways a parent-selector can go wrong in VRP (e.g., over-exploitation, diversity collapse, cost-feasibility trade-offs).
    \item \textbf{Strategies ($K_s$):} For each pitfall, propose one high-level countermeasure (e.g., partial elitism + random injection).
    \item \textbf{Hidden Traps ($K_t$):} What tricky instance features (e.g., clustered deliveries, time-windows) could mislead selection?
\end{enumerate}

\textbf{Overall Objective:}
Only write a Python function \texttt{select\_parents} implementing a \emph{novel} parent selection strategy for a genetic algorithm (Hybrid Genetic Search) solving the Vehicle Routing Problem (VRP). Unlike standard strategies, your goal is to propose a \emph{new} method—do not replicate existing implementations exactly. The function will be used within the PyVRP library. \textbf{Use your planning insights} to guide design. Do not redefine any of the provided classes; just import them if needed. Lower cost is better.
\end{tcolorbox}

\begin{tcolorbox}[colback=gray!5, colframe=gray!50, title=Context and Existing Approaches]
\footnotesize

\textbf{Context:}
In Hybrid Genetic Search for VRP, parents are selected from a population of solutions to undergo crossover. Parent selection directly influences exploration (diversity) and exploitation (quality) in the search. Standard selection strategies include:
\begin{itemize}
    \item Fitness-based selection
    \item Tournament selection  
    \item Random selection with diversity bias
\end{itemize}

\textbf{Common Existing Approaches:}
\begin{itemize}
    \item \textbf{Elitist Selection:} Always select top-ranked (lowest-cost) individuals.
    \item \textbf{Tournament Selection:} Randomly sample a subset of the population and pick the best.
    \item \textbf{Fitness Proportionate Selection:} Choose solutions with probability proportional to inverse cost.
    \item \textbf{Diversity-Aware Selection:} Combine one elite solution with one structurally different (e.g., high Hamming distance) solution.
    \item \textbf{Random Selection:} Uniform random sampling from feasible individuals.
\end{itemize}
\end{tcolorbox}

\begin{tcolorbox}[colback=gray!5, colframe=gray!50, title=Reasoning Phase and Task Description]
\footnotesize

\textbf{\textsc{2) Reasoning Phase}}

\emph{Now dig into \texttt{sel1} and \texttt{sel2}—but think aloud.}
\begin{itemize}
    \item \textbf{Identify Patterns:} What exploration/exploitation bias does each show?
    \item \textbf{Assess Shortcomings:} Where does each stumble relative to your $K_p$/$K_t$ list?
    \item \textbf{Design Sketch:} Using your $K_s$ strategies, outline in prose how \texttt{sel3} will avoid those failures.
\end{itemize}

\textbf{Your Task in This Interaction:}
You will be presented with two existing implementations of parent selection logic which are in Python, \texttt{sel1} and \texttt{sel2}, along with observed performance metrics. Your goal is to propose a \emph{new} Python implementation, \texttt{sel3}, that performs better (produces better children in crossover) and less cost compared to the given ones.
\end{tcolorbox}

\begin{tcolorbox}[colback=gray!5, colframe=gray!50, title=Implementation Instructions]
\footnotesize

\textbf{Instructions:}

\begin{enumerate}
    \item \textbf{Analyze \texttt{sel1} and \texttt{sel2}:}
    \begin{itemize}
        \item Understand how it balances exploration and exploitation.
        \item Evaluate whether it favors elite, random, or diverse solutions—and if it filters infeasible ones.
        \item Identify reasons why it may not generalize or may produce low-quality parents.
    \end{itemize}

    \item \textbf{Design \texttt{sel3 = select\_parents(population, rng, cost\_evaluator, k=2)}:}
    \begin{itemize}
        \item Propose a novel selection mechanism that improves the expected quality and diversity of selected parents.
        \item Consider hybrid strategies (e.g., partial elitism, feasibility filters, diversity-aware tournaments).
        \item Avoid trivial combinations of existing methods.
    \end{itemize}
\end{enumerate}
\end{tcolorbox}

\begin{tcolorbox}[colback=gray!5, colframe=gray!50, title=Implementation Requirements]
\footnotesize

\begin{enumerate}
\setcounter{enumi}{2}
    \item \textbf{Implementation Requirements:}
    \begin{itemize}
        \item \textbf{Input:} 
        \begin{itemize}
            \item \texttt{population: list[Solution]}
            \item \texttt{rng: RandomNumberGenerator}
            \item \texttt{cost\_evaluator: CostEvaluator}
            \item \texttt{k: int} (number of parents, always 2)
        \end{itemize}
        \item \textbf{Output:} \texttt{tuple[Solution, Solution]}
        \item \textbf{Function Signature:} Must be exactly:
        \begin{lstlisting}[language=Python, basicstyle=\ttfamily\tiny]
def select_parents(
    population: list[Solution],
    rng: RandomNumberGenerator,
    cost_evaluator: CostEvaluator,
    k: int = 2
) -> tuple[Solution, Solution]:
        \end{lstlisting}
        \item \textbf{Behavior:}
        \begin{itemize}
            \item Prioritize feasible solutions but allow occasional infeasible ones if beneficial.
            \item Use cost\_evaluator.penalised\_cost(solution) to assess solution quality.
            \item Ensure diversity between selected parents.
            \item Avoid modifying the population in-place.
        \end{itemize}
        \item \textbf{Performance Hints:}
        \begin{itemize}
            \item Cache penalized costs if reused multiple times.
            \item Minimize sorting or scanning the entire population.
            \item Use \texttt{rng.randint(len(population))} to sample indices efficiently.
        \end{itemize}
        \item \textbf{Constraints:}
        \begin{itemize}
            \item Use only Python built-ins and PyVRP classes.
        \end{itemize}
    \end{itemize}
\end{enumerate}
\end{tcolorbox}

\begin{tcolorbox}[colback=gray!5, colframe=gray!50, title=Existing Parent Selectors]
\footnotesize

\textbf{Pair of Existing Parent Selectors:}

\textbf{Selector 1:}
\begin{lstlisting}[language=Python, basicstyle=\ttfamily\tiny]
$code1
\end{lstlisting}
Current Cost: \texttt{\$score1}

Cost to beat (Target Cost/Current best cost): \texttt{\$baseline\_score}

Feedback: \texttt{\$feedback1}

\textbf{Selector 2:}
\begin{lstlisting}[language=Python, basicstyle=\ttfamily\tiny]
$code2
\end{lstlisting}
Current Cost: \texttt{\$score2}

Cost to beat (Target Cost/Current best cost): \texttt{\$baseline\_score}

Feedback: \texttt{\$feedback2}
\end{tcolorbox}

\begin{tcolorbox}[colback=gray!5, colframe=gray!50, title=Reference Classes]
\footnotesize

\textbf{Reference Python Classes Implementation:}
\textbf{The following class definitions are already available in another file and are provided here for reference only. Please do not redefine them. They can be imported using: \texttt{from pyvrp.\_pyvrp import CostEvaluator, RandomNumberGenerator, Solution}, etc.}

\begin{lstlisting}[language=Python, basicstyle=\ttfamily\tiny, breaklines=true]
class CostEvaluator:
    def __init__(
        self,
        load_penalties: list[float],
        tw_penalty: float,
        dist_penalty: float,
    ) -> None: ...
    def load_penalty(
        self, load: int, capacity: int, dimension: int
    ) -> int: ...
    def tw_penalty(self, time_warp: int) -> int: ...
    def dist_penalty(self, distance: int, max_distance: int) -> int: ...
    def penalised_cost(self, solution: Solution) -> int: ...
    def cost(self, solution: Solution) -> int: ...

class Solution:
    def __init__(
        self,
        data: ProblemData,
        routes: list[Route] | list[list[int]],
    ) -> None: ...
    @classmethod
    def make_random(
        cls, data: ProblemData, rng: RandomNumberGenerator
    ) -> Solution: ...
    def is_feasible(self) -> bool: ...
    # ... (additional methods omitted for brevity)

class RandomNumberGenerator:
    @overload
    def __init__(self, seed: int) -> None: ...
    @overload
    def __init__(self, state: list[int]) -> None: ...
    def rand(self) -> float: ...
    def randint(self, high: int) -> int: ...
    # ... (additional methods omitted for brevity)

# ... (other PyVRP class definitions omitted for brevity)
\end{lstlisting}
\end{tcolorbox}

\begin{tcolorbox}[colback=gray!5, colframe=gray!50, title=Reflection Phase and Response Format]
\footnotesize

\textbf{\textsc{3) Reflection Phase}}

\emph{After you draft \texttt{sel3}, self-evaluate before returning code.}
\begin{enumerate}
    \item Which of your original $K_p$ pitfalls does the final code best address? Which remain?
    \item Did any new traps surface during coding? How would you tweak \texttt{sel3} to handle them?
    \item Propose \textbf{two} concrete revisions to improve quality/diversity trade-off further.
\end{enumerate}

\textbf{Response Format:}
Enclose your answer in a single Python code block (\texttt{python}) with only the function \texttt{select\_parents} and necessary imports. The code must be directly extractable.

\textbf{Response Format Code:}
\begin{lstlisting}[language=Python, basicstyle=\ttfamily\tiny]
from pyvrp._pyvrp import CostEvaluator, RandomNumberGenerator, Solution

def select_parents(
    population: list[Solution],
    rng: RandomNumberGenerator,
    cost_evaluator: CostEvaluator,
    k: int = 2
) -> tuple[Solution, Solution]:
    """
    [Brief description of your novel parent selection strategy]
    """
    # [Your implementation here]
    parent1 = ...
    parent2 = ...
    return parent1, parent2
\end{lstlisting}
\end{tcolorbox}




\section{Parent Selection Code}
\label{append:parent_selection_code}

This section presents the complete implementation, from Figures \ref{fig:ohasdpps_algorithm_1} to \ref{fig:ohasdpps_algorithm_4}, of the parent selection algorithm that was discovered. The code is divided to ensure related functionality stays together while keeping each chunk at a reasonable size for page layout.

\begin{figure*}[!htbp]
\centering
\begin{minipage}{\textwidth}
\footnotesize
\begin{lstlisting}[language=Python, basicstyle=\ttfamily\scriptsize, 
                   breaklines=true, breakatwhitespace=true,
                   showstringspaces=false, frame=single,
                   numbers=left, numberstyle=\tiny,
                   commentstyle=\color{gray}\itshape,
                   keywordstyle=\color{blue}\bfseries,
                   stringstyle=\color{red}]
from pyvrp._pyvrp import CostEvaluator, RandomNumberGenerator, Solution

def select_parents(population: list[Solution], rng: RandomNumberGenerator, 
                  cost_evaluator: CostEvaluator, k: int=2) -> tuple[Solution, Solution]:
    """
    Hybrid Adaptive Stratified Diversity-Pressure Parent Selector
    -----------------------------------------------------------------------------------

    Summary:
    This selector provides scalable parent selection that maintains structural diversity awareness
    for manageable populations while gracefully degrading to efficient cost-based selection
    for large-scale problems. It employs population-size adaptive strategies to balance quality
    and computational efficiency, preventing timeouts in large populations while preserving
    diversity-driven selection for smaller populations.

    Step-by-step:
    ---------------
    1. **Precompute Costs & Stratification**:
        - Cache penalized costs and feasibility flags for all solutions upfront.
        - Sort population by cost and partition into three strata:
          * Elite: top 1/6 solutions (highest quality)
          * Mid: solutions from 1/6 to 2/3 (moderate quality)
          * Tail: bottom 1/3 solutions (lower quality)

    2. **Parent 1 Selection**:
        - Select from elite stratum using feasibility-cost biased tournament.
        - Tournament emphasizes best feasible solutions with occasional infeasible survivors
          (20% probability) for diversity maintenance.

    3. **Adaptive Parent 2 Strategy**:
        - Determine complementary strata based on Parent 1 characteristics:
          * If Parent 1 is elite & feasible : select from mid+tail (exploration)
          * If Parent 1 is infeasible or non-elite : select from elite+mid (intensification)

    4. **Population-Size Adaptive Selection**:
        - **Large populations (>500)**: Skip structural analysis entirely, use simplified scoring:
          * Cost advantage (60% weight)
          * Feasibility bonus (30% weight) 
          * Basic diversity bonus (10% weight)
        - **Small populations (=<500)**: Full structural analysis with lazy encoding:
          * Structural dissimilarity via client-set overlap (55% weight)
          * Cost advantage (30% weight)
          * Feasibility bonus (15% weight)

    5. **Lazy Encoding & Sampling**:
        - Only encode solutions that are actually evaluated as candidates.
        - Adaptive sampling limits: 20-30 candidates per stratum depending on population size.
        - Add small randomness to scores to avoid local minima.

    -----------------------------------------------------------------------------------
    This approach explicitly counters:
      - Computational timeouts in large populations through adaptive complexity reduction.
      - Over-exploitation by maintaining structural diversity awareness when computationally feasible.
      - Population stratification collapse through complementary strata selection strategy.
      - Unnecessary computational overhead by employing lazy encoding and adaptive sampling.
      - Pure cost or pure diversity bias through weighted multi-criteria scoring.

    Implementation is efficient: 
    Costs and feasibility are cached once; structural encoding is lazy and only applied 
    when needed; population-size thresholds automatically switch between full analysis 
    and simplified selection; adaptive sampling prevents excessive candidate evaluation.
    """
    
    n = len(population)
    assert n >= 2, 'Population must have at least two solutions.'
    assert k == 2, 'Only k=2 supported.'
    
    # Cache costs and feasibility upfront for efficient repeated access
    costs = [cost_evaluator.penalised_cost(sol) for sol in population]
    feasible = [sol.is_feasible() for sol in population]
    sorted_indices = sorted(range(n), key=lambda i: costs[i])
    
    # Partition population into three strata by cost quantiles
    # Elite: top 1/6, Mid: 1/6 to 2/3, Tail: bottom 1/3
    elite_cut = max(1, n // 6)
    mid_cut = max(elite_cut + 1, n * 2 // 3)
    elite_indices = sorted_indices[:elite_cut]
    mid_indices = sorted_indices[elite_cut:mid_cut]
    tail_indices = sorted_indices[mid_cut:]
    
\end{lstlisting}
\end{minipage}
\caption{Hybrid Adaptive Stratified Diversity-Pressure Parent Selector.}
\label{fig:ohasdpps_algorithm_1}
\end{figure*}

\begin{figure*}[!htbp]
\centering
\begin{minipage}{\textwidth}
\footnotesize
\begin{lstlisting}[language=Python, basicstyle=\ttfamily\scriptsize, 
                   breaklines=true, breakatwhitespace=true,
                   showstringspaces=false, frame=single,
                   numbers=left, numberstyle=\tiny,
                   commentstyle=\color{gray}\itshape,
                   keywordstyle=\color{blue}\bfseries,
                   stringstyle=\color{red}]
    def encode_solution_simple(solution: Solution):
        """
        Lightweight client-set based structural encoding for performance.
        Captures global client distribution and per-route client assignments
        without expensive edge-pair analysis from original version.
        """
        all_clients = set()
        route_clients = []
        for route in solution.routes():
            clients_in_route = set(v for v in route if v > 0)
            route_clients.append(clients_in_route)
            all_clients.update(clients_in_route)
        return (all_clients, route_clients)

    def simple_structural_distance(enc1, enc2):
        """
        Simplified structural dissimilarity based on client overlap patterns.
        Combines global client set similarity with route-level client distribution,
        replacing the complex harmonic mean of node/edge Jaccard distances.
        """
        all_clients1, route_clients1 = enc1
        all_clients2, route_clients2 = enc2
        
        # Global client set similarity (replacement for node-set similarity)
        if not all_clients1 and not all_clients2:
            return 0.0
        
        intersection = len(all_clients1 & all_clients2)
        union = len(all_clients1 | all_clients2)
        
        if union == 0:
            return 0.0
        
        global_similarity = intersection / union

        # Route structure similarity (simplified replacement for edge-set analysis)
        min_routes = min(len(route_clients1), len(route_clients2))
        if min_routes == 0:
            return 1.0 - global_similarity
        
        route_overlaps = []
        for i in range(min_routes):
            if not route_clients1[i] and not route_clients2[i]:
                continue
            r_intersection = len(route_clients1[i] & route_clients2[i])
            r_union = len(route_clients1[i] | route_clients2[i])
            if r_union > 0:
                route_overlaps.append(r_intersection / r_union)
        
        if route_overlaps:
            avg_route_similarity = sum(route_overlaps) / len(route_overlaps)
        else:
            avg_route_similarity = 0.0
        
        # Combine global and route-level similarity (70% global, 30% route-level weighting)
        combined_similarity = 0.7 * global_similarity + 0.3 * avg_route_similarity
        return 1.0 - combined_similarity
        
\end{lstlisting}
\end{minipage}
\caption{Hybrid Adaptive Stratified Diversity-Pressure Parent Selector.}
\label{fig:ohasdpps_algorithm_2}
\end{figure*}

\begin{figure*}[!htbp]
\centering
\begin{minipage}{\textwidth}
\footnotesize
\begin{lstlisting}[language=Python, basicstyle=\ttfamily\scriptsize, 
                   breaklines=true, breakatwhitespace=true,
                   showstringspaces=false, frame=single,
                   numbers=left, numberstyle=\tiny,
                   commentstyle=\color{gray}\itshape,
                   keywordstyle=\color{blue}\bfseries,
                   stringstyle=\color{red}]
    def tournament_select(indices, t_size=7, allow_infeasible_prob=0.2):
        """
        Combined feasibility-cost biased tournament emphasizing best feasible solutions
        with occasional infeasible survivors for diversity maintenance.
        """
        if len(indices) <= t_size:
            candidates = indices[:]
        else:
            candidates = [indices[rng.randint(len(indices))] for _ in range(t_size)]
        
        # Filter candidates by feasibility with probabilistic infeasible inclusion
        filtered = []
        for idx in candidates:
            if feasible[idx]:
                filtered.append(idx)
            elif rng.rand() < allow_infeasible_prob:
                filtered.append(idx)
        
        if not filtered:
            filtered = candidates
        
        best_idx = min(filtered, key=lambda i: costs[i])
        return best_idx

    # Select parent1 from elite stratum using feasibility-cost biased tournament
    parent1_idx = tournament_select(elite_indices)
    parent1 = population[parent1_idx]
    parent1_feas = feasible[parent1_idx]
    p1_cost_rank = sorted_indices.index(parent1_idx)
    
    # Determine complementary parent2 strata based on parent1 characteristics
    if parent1_feas and p1_cost_rank < elite_cut:
        parent2_strata = mid_indices + tail_indices
    else:
        parent2_strata = elite_indices + mid_indices
    
    if not parent2_strata:
        parent2_strata = sorted_indices
    
    # Adaptive sampling limits to balance quality vs. performance
    # Reduce sample size for large populations to prevent timeout
    max_sample_size = min(20, len(parent2_strata)) if n > 500 else min(30, len(parent2_strata))
    
    # Sample parent2 candidates with collision avoidance
    sampled_parent2_indices = []
    seen = set()
    attempts = 0
    while len(sampled_parent2_indices) < max_sample_size and attempts < max_sample_size * 2:
        candidate = parent2_strata[rng.randint(len(parent2_strata))]
        if candidate != parent1_idx and candidate not in seen:
            sampled_parent2_indices.append(candidate)
            seen.add(candidate)
        attempts += 1
    
    # Fallback candidate selection if sampling fails
    if not sampled_parent2_indices:
        for idx in parent2_strata:
            if idx != parent1_idx:
                sampled_parent2_indices.append(idx)
                break
        else:
            # Last resort fallback
            sampled_parent2_indices.append(parent2_strata[0])
        
\end{lstlisting}
\end{minipage}
\caption{Hybrid Adaptive Stratified Diversity-Pressure Parent Selector.}
\label{fig:ohasdpps_algorithm_3}
\end{figure*}

\begin{figure*}[!htbp]
\centering
\begin{minipage}{\textwidth}
\footnotesize
\begin{lstlisting}[language=Python, basicstyle=\ttfamily\scriptsize, 
                   breaklines=true, breakatwhitespace=true,
                   showstringspaces=false, frame=single,
                   numbers=left, numberstyle=\tiny,
                   commentstyle=\color{gray}\itshape,
                   keywordstyle=\color{blue}\bfseries,
                   stringstyle=\color{red}]
    # Population-size adaptive parent2 selection strategy
    if n > 500:
        # Large population: Simplified selection focusing on cost and feasibility
        # Skip structural analysis entirely to prevent timeouts
        best_score = -float('inf')
        parent2_idx = None
        
        p2_costs = [costs[i] for i in sampled_parent2_indices]
        p2_min_cost = min(p2_costs)
        p2_max_cost = max(p2_costs) if max(p2_costs) > p2_min_cost else p2_min_cost + 1
        
        for idx in sampled_parent2_indices:
            if idx == parent1_idx:
                continue
            
            cand_cost = costs[idx]
            cand_feas = feasible[idx]
            
            # Simplified three-component scoring without structural analysis
            cost_score = 1.0 - (cand_cost - p2_min_cost) / (p2_max_cost - p2_min_cost)
            feas_bonus = 1.0 if cand_feas else 0.0
            diversity_bonus = 0.5 if abs(costs[idx] - costs[parent1_idx]) > (p2_max_cost - p2_min_cost) * 0.1 else 0.0
            
            # Weighted combination emphasizing cost with feasibility and basic diversity
            score = 0.6 * cost_score + 0.3 * feas_bonus + 0.1 * diversity_bonus
            score += (rng.rand() - 0.5) * 0.02  # Small randomness to avoid local minima
            
            if score > best_score:
                best_score = score
                parent2_idx = idx

    else:
        # Small population: Full structural analysis with lazy encoding
        # Only encode parent1 and actual candidates (not entire population)
        p1_enc = encode_solution_simple(population[parent1_idx])
        
        p2_costs = [costs[i] for i in sampled_parent2_indices]
        p2_min_cost = min(p2_costs)
        p2_max_cost = max(p2_costs) if max(p2_costs) > p2_min_cost else p2_min_cost + 1
        
        # Weighted combination of three normalized signals
        w_struct = 0.55  # Structural dissimilarity weight
        w_cost = 0.3     # Cost advantage weight  
        w_feas = 0.15    # Feasibility bonus weight
        best_score = -float('inf')
        parent2_idx = None
        
        for idx in sampled_parent2_indices:
            if idx == parent1_idx:
                continue
            
            cand_cost = costs[idx]
            cand_feas = feasible[idx]
            cand_enc = encode_solution_simple(population[idx])  # Lazy encoding
            
            # Three-component scoring with structural dissimilarity
            struct_dist = simple_structural_distance(p1_enc, cand_enc)
            cost_score = 1.0 - (cand_cost - p2_min_cost) / (p2_max_cost - p2_min_cost)
            feas_bonus = 1.0 if cand_feas else 0.0
            
            # Weighted combination avoiding pure cost or pure diversity bias
            score = w_struct * struct_dist + w_cost * cost_score + w_feas * feas_bonus
            score += (rng.rand() - 0.5) * 0.02  # Small randomness to avoid local minima
            
            if score > best_score:
                best_score = score
                parent2_idx = idx
    
    # Final fallback if no suitable parent2 found
    if parent2_idx is None:
        parent2_idx = tournament_select(parent2_strata)
    
    parent2 = population[parent2_idx]
    return (parent1, parent2)

\end{lstlisting}
\end{minipage}
\caption{Hybrid Adaptive Stratified Diversity-Pressure Parent Selector.}
\label{fig:ohasdpps_algorithm_4}
\end{figure*}

\section{Supplementary Discussion}\label{append:supplementary}

\subsection{Comparison with Self-Refine and Self-Consistency}\label{append:selfrefine}

While approaches such as Self-Refine~\cite{madaan2023selfrefine} and Self-Consistency~\cite{wang2023selfconsistency} have demonstrated the value of iterative LLM self-improvement, they operate at fundamentally different levels of abstraction compared to MEP. Self-Refine enables an LLM to critique and polish a generated output through successive rounds of feedback, functioning as an output-level refinement loop. Self-Consistency samples multiple reasoning paths and selects the most frequent answer, improving reliability through redundancy. Neither approach enforces the explicit diagnostic reasoning that MEP mandates.

MEP draws specific inspiration from the metacognitive framework of Bai et al.~\cite{bai2025mp}, which models human-like planning and reasoning \emph{before} acting. The critical distinction lies in the structured Reason-Act-Reflect cycle: before generating any code, the LLM must (1) explicitly diagnose the weaknesses of parent heuristics against known failure modes ($K_p$, $K_t$), (2) formulate a concise, testable design hypothesis, and (3) reflect on its own implementation post-hoc. This proactive, hypothesis-driven process is qualitatively different from reactive refinement. In Self-Refine, the model generates first and critiques second; in MEP, diagnosis and hypothesis formulation necessarily precede implementation. This ordering is critical for complex algorithmic design tasks, where generating code without a clear strategic intent leads to superficial mutations rather than principled innovations. In the classical hyper-heuristic taxonomy, MEP thus functions as a metacognitive construction hyper-heuristic, operating in the space of heuristic generation rather than heuristic selection, with the LLM serving as both the generator and the strategic reasoner.

\subsection{Illustrative Failure-Reflection Case}\label{append:failure_case}

To concretize the Reason-Act-Reflect cycle, we describe a representative case observed during the evolution of the \texttt{select\_parents} operator.

In generation 3, the LLM was presented with two parent heuristics: one that selected parents purely by cost rank (achieving low average cost but exhibiting premature convergence on clustered instances), and another that used uniform random selection (maintaining diversity but producing poor-quality offspring).

\paragraph{Reason.} The LLM diagnosed that the first parent suffered from diversity collapse---a pitfall identified in $K_p$---noting that \emph{``selecting exclusively from the top cost stratum causes the population to converge to a narrow region of the solution space, particularly on instances with clustered customer distributions.''}  It further identified that the second parent's lack of quality pressure wasted computational budget on unpromising crossover pairs.

\paragraph{Act.} The LLM formulated the design hypothesis: \emph{``A stratified selection mechanism that draws the first parent from the elite stratum and the second parent from complementary strata, weighted by structural dissimilarity, will balance exploitation of high-quality solutions with exploration of diverse genetic material.''} This led to the implementation of cost-based stratification with adaptive diversity pressure.

\paragraph{Reflect.} The LLM self-critiqued that its implementation addressed the diversity collapse pitfall ($K_p$) but noted a potential vulnerability: \emph{``On very small populations ($n < 10$), the stratification boundaries may yield empty strata, requiring fallback logic.''} This reflection was carried forward and informed the next generation, which introduced explicit handling of degenerate population sizes.

This case illustrates how the structured cycle produces targeted improvements grounded in diagnosed weaknesses, rather than arbitrary code mutations.

\begin{table*}[h]
\centering
\caption{Performance comparison of MEP-Evolved, PyVRP Baseline, and POMO-MTL (DRL Solver) on the CVRP benchmark. Lower cost indicates better performance.}
\label{tab:neural_comparison}
\renewcommand\arraystretch{1.5}
\begin{tabular}{lcc}
\toprule
\textbf{Method} & \textbf{Average Cost} & \textbf{Gap vs.\ HGS Baseline} \\
\midrule
MEP-Evolved (Ours) & 64\,236.30 & $-$0.33\,\% (Improvement) \\
PyVRP Baseline (HGS) & 64\,450.03 & 0.00\,\% \\
POMO-MTL (DRL Solver) & 71\,575.70 & +11.06\,\% (Worse) \\
\bottomrule
\end{tabular}
\end{table*}

\subsection{Comparison with Neural Solvers}\label{append:neural_solver}

To contextualize the performance of MEP-evolved heuristics relative to neural combinatorial optimization approaches, we evaluated POMO-MTL, a representative deep reinforcement learning (DRL)-based solver, on the standard CVRP test set. The results are presented in Table~\ref{tab:neural_comparison}.

While MEP improves upon the SOTA HGS baseline, the DRL-based solver lags behind by a substantial margin of 11.06\%. This comparison underscores a key motivation of our work: rather than replacing traditional solvers with end-to-end neural policies---which currently cannot match the performance of well-engineered metaheuristics on realistic VRP instances---MEP enhances the existing SOTA solver from within, leveraging the LLM's reasoning capabilities to discover better algorithmic components while preserving the proven search framework of HGS.

\subsection{Detailed Comparison with ReEvo and EoH}\label{append:reevo_comparison}

The ``Reactive Evolution'' baseline in Table~\ref{tab:ablation_study} is explicitly designed to simulate the core mechanics of prior feedback-driven LLM-based hyper-heuristics, including ReEvo~\cite{ye2024reevo} and Evolution of Heuristics (EoH)~\cite{liu2024evolution}. In this variant, the prompt provides only the parent heuristics and their performance scores, asking the LLM to generate an improved version based on direct feedback, without any structured diagnosis, hypothesis formulation, or self-reflection.

The results reveal a stark contrast. On VRPTW, Reactive Evolution achieves only a 0.10\% improvement, while the full MEP framework achieves 2.25\% under identical conditions---a 22$\times$ difference in effectiveness. More critically, Reactive Evolution produces performance degradation on three of six variants ($-$0.20\% on CVRP, $-$0.39\% on VRPB, $-$2.05\% on MDVRPTW), indicating that feedback-driven mutation without structured reasoning can actively harm the solver when applied to complex algorithmic components. In contrast, the full MEP framework achieves positive improvements on all six variants without exception.

These results suggest that while the reactive paradigm of ReEvo and EoH is effective for evolving simpler constructive heuristics for basic problem variants, it is insufficient for the more challenging task of improving components within a sophisticated, well-optimized SOTA solver. The metacognitive scaffolding of MEP---specifically the mandate to diagnose failures and formulate hypotheses before code generation---is the critical ingredient that enables effective heuristic discovery at this level of algorithmic complexity.

\newpage
\subsection{Limitations}\label{append:limitations}

We acknowledge several limitations of the current work.

\paragraph{Proprietary Model Dependence.} Our framework relies on GPT-4.1 for heuristic generation. While our ablation study (Table~\ref{tab:ablation_study}) demonstrates that it is the MEP methodology---not the specific LLM---that drives performance gains (as evidenced by the clear hierarchy between Full MEP, MEP-noInit, and Reactive Evolution under the same model), the dependence on a proprietary API limits reproducibility and accessibility. In preliminary experiments, smaller open-source models (e.g., LLaMA-70B) struggled with the complex reasoning and strict syntactic requirements needed for valid Python code generation in this context. As open-source models continue to mature in code generation and structured reasoning capabilities, adapting MEP to leverage these models is a priority for future work.

\paragraph{Evolutionary Budget.} Budget constraints limited the evolutionary process to $N{=}10$ generations. While Figure~\ref{fig:fitness_evolution} demonstrates rapid convergence within this limit, a more extensive search with larger generation budgets could potentially discover even more effective heuristics. The current results nonetheless provide strong evidence that MEP efficiently locates high-quality regions of the search space with a manageable one-time investment.

\paragraph{Generalization Scope.} While the evolved heuristics generalize well across the six VRP variants evaluated---despite being evolved exclusively on TSP100 instances---further validation across a broader set of problem distributions, larger instance scales, and the full set of 48 VRP variants supported by PyVRP is needed to fully characterize the boundaries of this generalization.

\end{document}